\documentclass[a4paper,conference]{IEEEtran}
\usepackage{graphicx}
\usepackage{amsmath}
\usepackage{amssymb}
\usepackage{booktabs}
\usepackage{array}
\usepackage{multirow}
\usepackage{color}
\usepackage{colortbl}
\usepackage{framed}
\usepackage{bm}
\usepackage{bbm}
\usepackage{xspace}
\usepackage{enumitem}
\usepackage[dvipsnames,table]{xcolor}
\usepackage{pifont}
\usepackage{scalerel}
\usepackage{xparse}
\usepackage{xcolor}
\usepackage{algorithm}
\usepackage{lipsum}
\usepackage{listings}
\usepackage{url}

\usepackage[pagebackref,breaklinks]{hyperref}

\def \ie {\emph{i.e.}}

\def \etal {\emph{et al.}}

\xspaceaddexceptions{\textsubscript}
\newcommand{\ours}{CaMEL\xspace}

\newcommand{\cmark}{\ding{51}}

\newcommand{\tit}[1]{\smallbreak\noindent\textbf{#1.}}
\newcommand{\tinytit}[1]{\noindent\textbf{#1.}}

\begin{document}
%
\title{CaMEL: Mean Teacher Learning for\\Image Captioning}

\author{\IEEEauthorblockN{Manuele Barraco, Matteo Stefanini, Marcella Cornia, Silvia Cascianelli, Lorenzo Baraldi, Rita Cucchiara}
\IEEEauthorblockA{University of Modena and Reggio Emilia\\
Email: \{name.surname\}@unimore.it}
}

\maketitle

\begin{abstract}
Describing images in natural language is a fundamental step towards the automatic modeling of connections between the visual and textual modalities. In this paper we present CaMEL, a novel Transformer-based architecture for image captioning. Our proposed approach leverages the interaction of two interconnected language models that learn from each other during the training phase. The interplay between the two language models follows a mean teacher learning paradigm with knowledge distillation.  Experimentally, we assess the effectiveness of the proposed solution on the COCO dataset and in conjunction with different visual feature extractors. When comparing with existing proposals, we demonstrate that our model provides state-of-the-art caption quality with a significantly reduced number of parameters. According to the CIDEr metric, we obtain a new state of the art on COCO when training without using external data. The source code and trained models are publicly available at: \url{https://github.com/aimagelab/camel}.
\end{abstract}


%
\IEEEpeerreviewmaketitle

\section{Introduction}
\label{sec:introduction}
Image captioning has received a relevant interest in the last few years, as describing images in natural language is a fundamental step to model the interconnections between the visual and textual modalities~\cite{stefanini2022show}. Early works on this research line were based on the usage of convolutional neural networks~\cite{karpathy2015deep,vinyals2015show,xu2015show,vinyals2016show} -- usually pre-trained to solve classification tasks -- or object detectors~\cite{anderson2018bottom,huang2019attention,liu2020prophet} as visual feature extractors, and recurrent neural networks as auto-regressive language models~\cite{karpathy2015deep,vinyals2015show,anderson2018bottom,rennie2017self,landi2021working}.
Nowadays, captioning approaches have significantly evolved towards the usage of Transformer-based language models~\cite{herdade2019image,pan2020x,cornia2020meshed,luo2021dual,cornia2021explaining}, while the visual feature extraction stage is rapidly evolving towards the use of grid-like features extracted with multi-modal architectures trained on large-scale data with language supervision~\cite{shen2021much,cornia2021universal,radford2021learning}. In parallel, while many previous captioning approaches have been trained on middle-size datasets like COCO, the literature is now investigating the usage of large-scale noisy datasets as well~\cite{cornia2021universal,li2020oscar,zhang2021vinvl,wang2021simvlm,hu2021scaling}.

\begin{figure}[t]
\centering
\includegraphics[width=\linewidth]{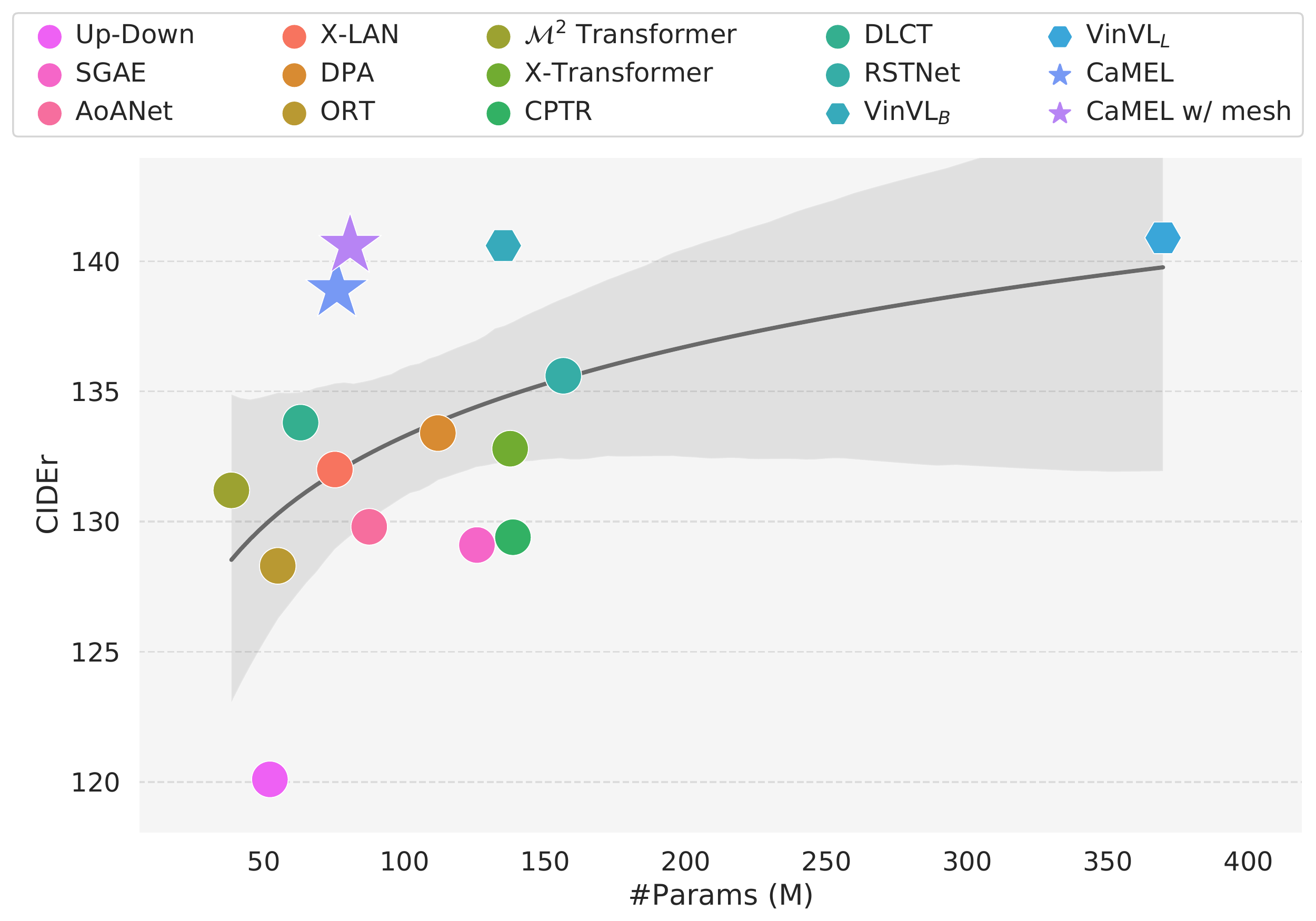}
\caption{Comparison of two different versions of our approach (marked with stars) and existing approaches (marked with bullets, and hexagons if exploiting vision-and-language pre-training) in terms of number of parameters and caption quality. Our method features state-of-the-art caption quality in terms of CIDEr with a significantly reduced number of parameters.}
\label{fig:first_page}
\vspace{-0.15cm}
\end{figure}

Regardless of these architectural and structural improvements, the training methodology has remained almost unaltered. Indeed, most of the existing approaches for image captioning are based on the usage of a single language model, trained to reproduce the ground-truth caption through a cross-entropy loss and later, in a fine-tuning stage, through the REINFORCE algorithm~\cite{rennie2017self,liu2017improved}. In this paper, we take a different path and investigate the development of a training strategy that is based on the interplay between two distinct language models. In particular, we draw inspiration from the Mean Teacher Learning approach~\cite{tarvainen2017mean} -- which has been successfully employed to learn visual representation in a self-supervised manner~\cite{caron2021emerging} -- and propose a schema in which two language models learn and interact together at training time. One of the two language models is employed as teacher, while the other is employed as a student in a knowledge distillation relationship~\cite{hinton2015distilling,anil2018large,xie2020self}. Parameters update, on the teacher model, is carried out by averaging successive states of the student, through an exponential moving average. In this way, the teacher slowly ``follows'' the student state through time. We devise and compare strategies to apply this interplay paradigm in both the cross-entropy training stage and during the fine-tuning with reinforcement learning.

Noticeably, at test time one of the two language models can be discarded, so that the number of parameters is kept on pair with traditional models that employ a single language model at training time. We name our model \ours~-- short for Captioner with Mean tEacher Learning. As shown in Fig.~\ref{fig:first_page}, our model outperforms existing approaches in terms of caption quality, while being significantly less demanding in terms of number of parameters. We assess the performances of the proposed training strategy on the COCO dataset~\cite{lin2014microsoft}, employing different knowledge distillation strategies,  and in comparison with other state-of-the-art approaches that have been trained on the same dataset. We also compare our results on the COCO online test server. Results demonstrate the appropriateness of the proposed solution, which attains a new state of the art on COCO when training without using external data.

\section{Related Work}
\label{sec:related}
\begin{figure*}
    \centering
    \includegraphics[width=.98\textwidth]{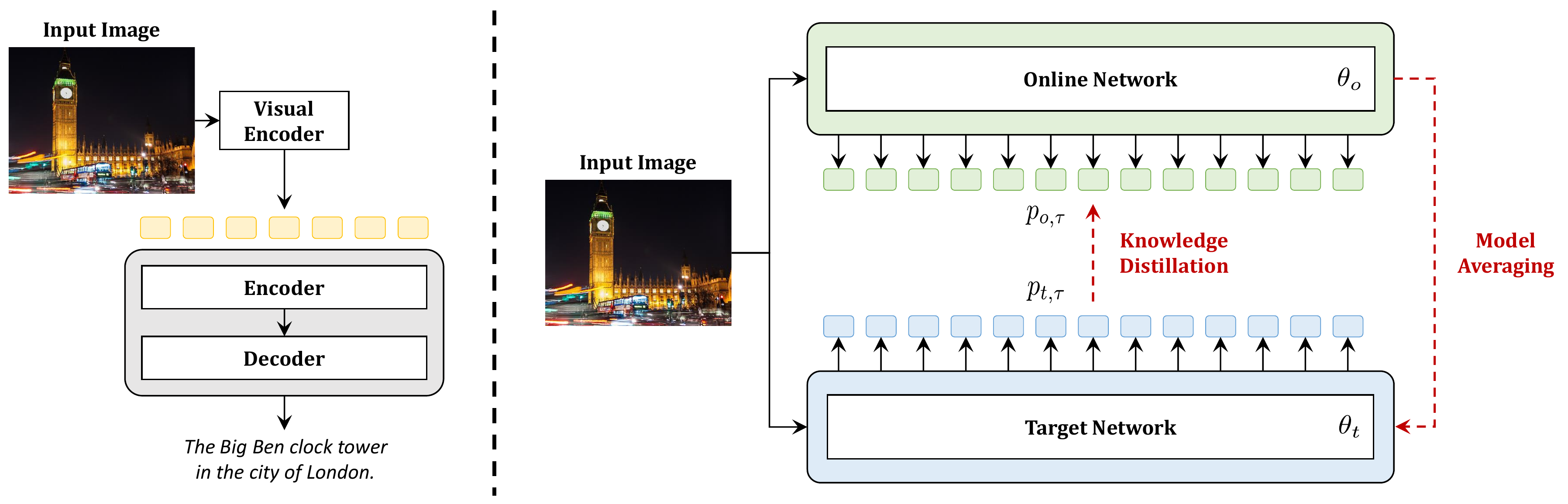}
    \caption{Overview of our approach and of the interplay between the online and target language models.}
    \label{fig:overview}
\end{figure*}

Early deep learning-based image captioning approaches entail encoding the visual information by using a CNN-based encoder, and then generating the textual output by using RNN-based language models conditioned on the encoding~\cite{karpathy2015deep,vinyals2015show,rennie2017self,donahue2015long}. 

\tit{Visual encoding} To better capture spatial information and structure, the encoding component has been modified to introduce attention mechanisms, which rely on grids of CNN features in early works~\cite{xu2015show,lu2017knowing,you2016image}, and on image regions containing visual entities in following works~\cite{anderson2018bottom,lu2018neural}. An alternative strategy to model spatial and semantic relations between the objects in the image is employing graph convolution neural networks~\cite{yao2018exploring,yang2019auto,shi2020improving}. 
Finally, a more recent trend entails applying Transformer-like architectures as visual encoders~\cite{cornia2021explaining,liu2021cptr}, being those also applicable to image patches directly~\cite{dosovitskiy2021image,touvron2021training}. In our case, we follow the recent paradigm of employing features extracted from large-scale multi-modal architectures~\cite{shen2021much,cornia2021universal} like CLIP~\cite{radford2021learning}.

\tit{Language model} Despite RNN-based language models have been the standard strategy for generating the caption, convolutional language models~\cite{aneja2018convolutional} and fully-attentive language models~\cite{luo2021dual,yang2019learning,li2019entangled,cagrandi2021learning,zhang2021rstnet} based on the Transformer paradigm~\cite{vaswani2017attention} have been explored for image captioning, also motivated by the success of these approaches on Natural Language Processing tasks such as machine translation and language understandings~\cite{vaswani2017attention,devlin2018bert,sukhbaatar2019augmenting}. Moreover, the introduction of Transformer-based language models has brought to the development of effective variants or modifications of the self-attention operator~\cite{huang2019attention,herdade2019image,pan2020x,cornia2020meshed,guo2020normalized,cornia2020smart,liu2020prophet} and has enabled vision-and-language early-fusion~\cite{li2020oscar,hu2021scaling,zhou2020unified}, based on BERT-like architectures~\cite{devlin2018bert}.

\tit{Training strategies}
The training strategy for image captioning architectures usually follows the time-wise cross-entropy paradigm. This was later combined with a fine-tuning phase based on the application of the REINFORCE algorithm, to allow using as optimization objectives captioning metrics directly~\cite{rennie2017self,liu2017improved}, overcoming the issue of their non-differentiability and boosting the final performance. 
As a strategy to improve both training phases, in~\cite{huang2020teacher} it is proposed to exploit a teacher model trained on image attributes to generate additional supervision signals for the captioning model. These are in the form of soft-labels, which the captioning model has to align with in the cross-entropy phase, and re-weighting of the caption words to guide the fine-tuning phase. Additional improvement to the performance of recent self-attention-based image captioning approaches is due to the use of large-scale vision-and-language pre-training~\cite{cornia2021universal,li2020oscar,zhang2021vinvl,hu2021scaling,zhou2020unified}, which can be done on noisy image-text pairs, also exploiting pre-training losses different from cross-entropy, such as the masked token loss~\cite{li2020oscar,zhang2021vinvl}. Different from previous methods, our approach is based on the interplay of two different language models that are trained with the mean teacher learning paradigm and knowledge distillation, without relying on large-scale pre-training.

\section{Approach}
\label{sec:approach}
\subsection{Preliminaries}
Most captioning approaches rely on a single language model, which is conditioned on input images and is trained to reproduce ground-truth sentences. Formally, given a dataset of image-caption pairs $\mathcal{D}=\{(\bm{v}_i, \bm{t}_i)\}_i$, the language model aims at learning the probability distribution of the next word in a sequence, conditioned on the input image, \ie
\begin{equation}
    p(\bm{w}_\tau | \bm{w}_{k < \tau}, \bm{v}),
\end{equation} 
where $\bm{v}$ is an input image, $\tau$ indicates time, and $\{\bm{w}_\tau\}_\tau$ is the sequence of words comprising the generated caption. The model is trained according to a time-wise cross-entropy (XE) loss over the entire dataset, as follows:
\begin{equation}
\mathcal{L}(\theta) = -\mathbb{E}_{\bm{x} \sim \mathcal{D}} \sum_\tau \log p(\bm{w}_\tau|\bm{w}_{k < \tau}, \bm{v}, \theta),
\end{equation}
where $\theta$ indicates the set of parameters of the model.

After a training stage with cross-entropy, sequence generation is usually fine-tuned using reinforcement learning. When training with XE, indeed, the model is trained to predict the next token given previous ground-truth words; in reinforcement learning the model is asked to generate an entire sequence and receives a reward that is proportional to the similarity of the generated caption with respect to the ground-truth. A standard practice is to employ a variant of the self-critical sequence training (SCST) approach~\cite{rennie2017self} on sequences sampled using beam search~\cite{anderson2018bottom}: to decode, the top-$k$ words are sampled from the language model probability distribution at each timestep, and a beam with the top-$k$ sequences with the highest probability is maintained during the generation.

Following previous works~\cite{anderson2018bottom}, the usual practice is to use the CIDEr-D score as reward, as it well correlates with human judgment~\cite{vedantam2015cider}. In our case, we baseline the reward using the mean of the rewards~\cite{cornia2020meshed}. 
The final gradient expression for the SCST training is, therefore
\begin{equation}
    \nabla_\theta \mathcal{L}(\theta) = -\frac{1}{k}\sum_{i=1}^k \left((r(\bm{w}^i)-b) \nabla_\theta \log p(\bm{w}^i)\right),
\end{equation}
where $\bm{w}^i$ is the $i$-th sentence in the beam, $r(\cdot)$ is the reward function, and $b = \left(\sum_i r(\bm{w}^i)\right)/k$ is the baseline, computed as the mean of the rewards obtained by the sampled sequences.

\subsection{\ours} 
In \ours, instead of training a single language model, we rely on the interplay of two different language models -- an \textit{online} and a \textit{target} language model, that interact and learn from each other during the training phase, both during the XE pre-training and during the SCST fine-tuning. At test time, each of the two language models can be used, alone, for captioning input images.

The interaction between the online and target language models at training time is two-fold. The online language model is trained, either via XE or SCST, with respect to ground-truth captions. In addition, it performs knowledge distillation with the target language model. The target language model, in turn, updates its weights according to an exponential moving average of the online model weights. An overview of our approach is given in Fig.~\ref{fig:overview}.

\tit{Knowledge distillation} The target network provides regression targets to train the online network. This is done through knowledge distillation -- treating the online language model as a student network and the target model as a teacher.

Given a visual input $\bm{v}$ and a conditioning partial sentence $\bm{w}$, at each timestep $\tau$ both networks provide output logits over a vocabulary of $N$ tokens, denoted as $p_{t,\tau}$ and $p_{o,\tau}$ for the target model and the online model, respectively. Given the teacher, which is kept fixed, we learn to match these distributions by minimizing a mean squared error loss with respect to the parameters of the online network, \ie
\begin{equation}
    \min_{\theta_o} \sum_\tau (p_{t,\tau} - p_{o,\tau})^2,
\end{equation}
where $\theta_o$ indicates the set of parameters of the online network.

\tit{Model averaging}  The parameters of the target language model are updated as an exponential moving average~\cite{caron2021emerging} of the parameters of the online network $\theta_o$. Formally, the parameters of the target model are given by 
\begin{equation}
    \theta_t \leftarrow \lambda \theta_t + (1-\lambda) \theta_o,
\end{equation}
where $\theta_t$ indicates the set of parameters of the target network and $\lambda \in \left[0,1\right]$ is a target decay rate. In practice, we keep $\lambda$ fixed during the entire training process. This strategy results in the target network keeping a weighted average of successive states from the online network, thus performing a form of model ensembling.

\tit{Objective} During XE pre-training, the final objective we employ to train the online network is a combination of the standard XE loss, which is computed with respect to ground-truth captions, and of the knowledge distillation loss with respect to the target logits. After each SGD update of the online network, the target network is updated through the model averaging. Algorithm~\ref{algo:xe_pseudocode} provides the PyTorch pseudo-code of the training loop during the XE stage.

\begin{algorithm}[tb]
   \caption{\ours PyTorch pseudocode}
    \definecolor{codeblue}{rgb}{0.25,0.5,0.5}
    \lstset{
      basicstyle=\fontsize{7.2pt}{7.2pt}\ttfamily\bfseries,
      commentstyle=\fontsize{7.2pt}{7.2pt}\color{codeblue},
      keywordstyle=\fontsize{7.2pt}{7.2pt},
    }
\label{algo:xe_pseudocode}
\begin{lstlisting}[language=python]
# gt, go: target and online networks
# l: network momentum
# v, c: training (image, caption) pair
for v, c in dataloader:
    t = gt(v, c)                     # target output
    o = go(v, c)                     # online output
	
    loss = XE(softmax(o, dim=-1), c) + MSE(t, o)
    loss.backward()                  # backpropagate
	
    update(go)                       # SGD
    gt.params = l*gt.params + (1-l)*go.params


def MSE(t, s):
    t = t.detach()                   # stop gradient
    return (t - s).square().mean()

\end{lstlisting}
\end{algorithm}

\tit{Extension to the SCST stage} The same training methodology is also applied during SCST fine-tuning. In this case, given that both language models generate a beam of $k$ captions, there are $k^2$ pairs of sequences on which the MSE loss can be potentially applied. In the following, we experiment by matching the top-1 caption in each beam, according to the probability assigned by the models themselves or to the score assigned by an external image-text model. Further, we also experiment when matching each caption in the online beam with a caption in the target beam, and then applying the MSE loss on each pair.

\begin{table*}[t]
\footnotesize
\centering
\caption{Results on the COCO Karpathy-test split with different visual encoders when training with cross-entropy loss.}
\label{tab:features}
\setlength{\tabcolsep}{.5em}
\resizebox{0.95\linewidth}{!}{
\begin{tabular}{lc ccc c cccccc c cccccc}
\toprule
& & & & & & \multicolumn{6}{c}{\textbf{Online Network}} & & \multicolumn{6}{c}{\textbf{Target Network}} \\
\cmidrule{7-12} \cmidrule{14-19}
 & & \ours & $\lambda_\text{KD}$ & w/ mesh & & B-1 & B-4 & M & R & C & S & & B-1 & B-4 & M & R & C & S\\
\midrule
\multirow{3}{*}{Faster R-CNN~\cite{anderson2018bottom,ren2017faster}} 
& & & - & & & - & - & - & - & - & - & & 75.4 & 35.8 & 27.9 & \textbf{56.4} & 113.9 & 20.7 \\
& & \cmark & 0.01 & & & 75.1 & 35.0 & 27.4 & 55.9 & 112.2 & 20.4 & & \textbf{75.8} & 36.0 & 27.9 & \textbf{56.4} & 114.7 & 20.6 \\
& & \cmark & 0.1 & & & 75.2 & 35.1 & 27.6 & 55.7 & 111.9 & 20.4 & & 75.7 & \textbf{36.1} & \textbf{28.0} & 56.3 & \textbf{114.8} & \textbf{20.8} \\
\midrule
\multirow{3}{*}{CLIP-RN50~\cite{radford2021learning}} 
& & & - & & & - & - & - & - & - & - & & 75.4 & 35.4 & 27.5 & 56.1 & 112.8 & 20.6 \\
& & \cmark & 0.01 & & & 74.1 & 34.1 & 27.6 & 55.4 & 110.4 & 20.5 & & 75.0 & 35.1 & 27.8 & 56.0 & 112.9 & 20.7 \\
& & \cmark & 0.1 & & & 75.5 & 35.4 & 27.6 & 56.3 & 113.7 & 20.6 & & \textbf{75.7} & \textbf{36.2} & \textbf{28.1} & \textbf{56.7} & \textbf{115.9} & \textbf{20.8} \\
\midrule
\multirow{3}{*}{CLIP-ViT-B16~\cite{radford2021learning}} 
& & & - & & & - & - & - & - & - & - & & 77.2 & 37.1 & 28.7 & 57.5 & 122.0 & 21.7 \\
& & \cmark & 0.01 & & & 77.0 & 36.8 & 28.6 & 57.5 & 119.6 & 21.5 & & 77.5 & 37.9 & \textbf{29.1} & \textbf{58.1} & \textbf{122.5} & \textbf{21.9} \\
& & \cmark & 0.1 & & & 76.6 & 36.3 & 28.3 & 56.9 & 117.9 & 21.2 & & \textbf{77.8} & \textbf{38.0} & \textbf{29.1} & 58.0 & \textbf{122.5} & 21.8 \\
\midrule
\multirow{6}{*}{CLIP-RN50$\times$16~\cite{radford2021learning}} 
& & & - & & & - & - & - & - & - & - & & 77.6 & 37.6 & 28.7 & 57.6 & 121.0 & 21.7 \\
& & \cmark & 0.01 & & & 78.0 & 37.4 & 28.7 & 57.8 & 121.4 & 21.9 & & 78.4 & 38.7 & 29.3 & \textbf{58.5} & 124.7 & \textbf{22.3} \\
& & \cmark & 0.05 & & & 77.2 & 37.4 & 29.0 & 57.9 & 121.4 & 21.9 & & 78.0 & 38.4 & \textbf{29.4} & \textbf{58.5} & 125.4 & 22.2 \\
& & \cmark & 0.1 & & & 77.7 & 37.8 & 29.0 & 58.0 & 122.3 & 22.0 & & 78.3 & \textbf{39.1} & \textbf{29.4} & \textbf{58.5} & \textbf{125.7} & 22.2 \\
& & \cmark & 0.5 & & & 77.9 & 37.6 & 28.5 & 57.7 & 121.1 & 21.6 & & \textbf{78.5} & 39.0 & 29.3 & \textbf{58.5} & 125.0 & 22.2 \\
& & \cmark & 1.0 & & & 77.7 & 36.7 & 28.3 & 57.3 & 120.0 & 21.7 & & \textbf{78.5} & 38.9 & 29.3 & \textbf{58.5} & 125.1 & \textbf{22.3} \\
\midrule
\multirow{6}{*}{CLIP-RN50$\times$16~\cite{radford2021learning}} 
& & & -  & \cmark & & - & - & - & - & - & - & & 77.5 & 37.6 & 29.0 & 58.0 & 122.6 & 21.9 \\
& & \cmark & 0.01 & \cmark & & 77.5 & 37.8 & 29.0 & 58.0 & 123.1 & 21.9 & & 78.0 & \textbf{38.8} & \textbf{29.4} & \textbf{58.6} & \textbf{125.0} & \textbf{22.2} \\
& & \cmark & 0.05 & \cmark & & 78.0 & 37.5 & 28.7 & 57.9 & 121.0 & 21.6 & & \textbf{78.2} & 38.5 & 29.3 & 58.4 & 124.4 & 22.1 \\
& & \cmark & 0.1 & \cmark & & 77.3 & 36.8 & 28.5 & 57.3 & 120.1 & 21.7 & & 78.0 & 38.5 & 29.3 & 58.4 & 124.2 & \textbf{22.2} \\
& & \cmark & 0.5 & \cmark & & 77.4 & 36.9 & 28.5 & 57.3 & 119.7 & 21.7 & & 77.9 & 38.6 & 29.3 & 58.3 & 124.2 & 22.1 \\
& & \cmark & 1.0 & \cmark & & 77.0 & 36.7 & 28.4 & 57.2 & 119.7 & 21.5 & & 77.7 & 38.3 & 29.3 & 58.2 & 123.7 & 22.0 \\
\bottomrule
\end{tabular}
}
\vspace{-0.15cm}
\end{table*}

\tit{Network architecture}
\ours follows an encoder-decoder Transformer~\cite{vaswani2017attention} architecture, where the encoder processes visual features via bi-directional attention and the decoder generates captions in an auto-regressive manner. Following~\cite{cornia2020meshed,cornia2020smart}, our encoder incorporates additional memory slots, enhancing its ability to encode knowledge and relations learned from visual data. Specifically, we expand the set of keys and values in self-attention layers with extra and independent learnable vectors, which can encode a priori knowledge retrieved through attention.
Our decoder is composed of a stack of decoder layers, each performing a right-masked self-attention and a cross-attention followed by a position-wise feed-forward network.

We also test the usage of a mesh-like connectivity between the encoder and the decoder, following~\cite{cornia2020meshed}. In this case, the mesh mechanism further connects each encoder and decoder layer in a mesh-like structure, augmenting its ability to deal with low- or high-level features.

The architecture is the same for both online and target models, while embracing independent parameters updated with different strategies during training.

\section{Experiments}
\label{sec:experiments}
\subsection{Dataset}
Following the dominant paradigm in literature, we train and evaluate our model on the COCO dataset~\cite{lin2014microsoft}. As such, we do not rely on large-scale image-text datasets~\cite{cornia2021universal}. COCO is composed of more than $120,000$ images, each of them associated with 5 human-collected captions.
We follow the Karpathy~\etal~\cite{karpathy2015deep} splits, using $5,000$ images for both validation and testing and the rest for training. We also evaluate our model on the COCO online test server, which includes $40,775$ images for which annotated captions are not publicly available.

\subsection{Experimental Settings}
\tinytit{Metrics} According to the standard evaluation protocol, we employ the complete set of captioning metrics: BLEU~\cite{papineni2002bleu}, METEOR~\cite{banerjee2005meteor}, ROUGE~\cite{lin2004rouge},  CIDEr~\cite{vedantam2015cider}, and SPICE~\cite{spice2016}.

\tit{Implementation details}
To represent words, we use Byte Pair Encoding (BPE)~\cite{sennrich2016neural} with a vocabulary size of $49,408$, which is then linearly projected to the input dimensionality of the model. We use standard sinusoidal positional encodings~\cite{vaswani2017attention} to represent word positions.
All models comprise three layers in the visual encoder and three layers in the decoder, each with a dimensionality of 512, a feed-forward dimensionality of 2048, and a number of heads equal to 8. 
We apply dropout at the output of each sub-layer, with a dropout probability equal to 0.1. The number of memory slots is set to 40.  

In all experiments, we employ Adam~\cite{kingma2015adam} as optimizer and a beam size equal to 5. During pre-training with XE loss, we use a batch size of 50, following the typical Transformer learning rate scheduling strategy~\cite{vaswani2017attention} with a warmup equal to $10,000$ iterations. During the SCST finetuning stage, we use a batch size of 30 and a fixed learning rate equal to $5\times10^{-6}$. 
The MSE loss is computed by considering only valid tokens and masking the rest. The target network momentum $\lambda$ is set to $0.999$.

\begin{figure*}[t]
\centering
\includegraphics[width=0.98\textwidth]{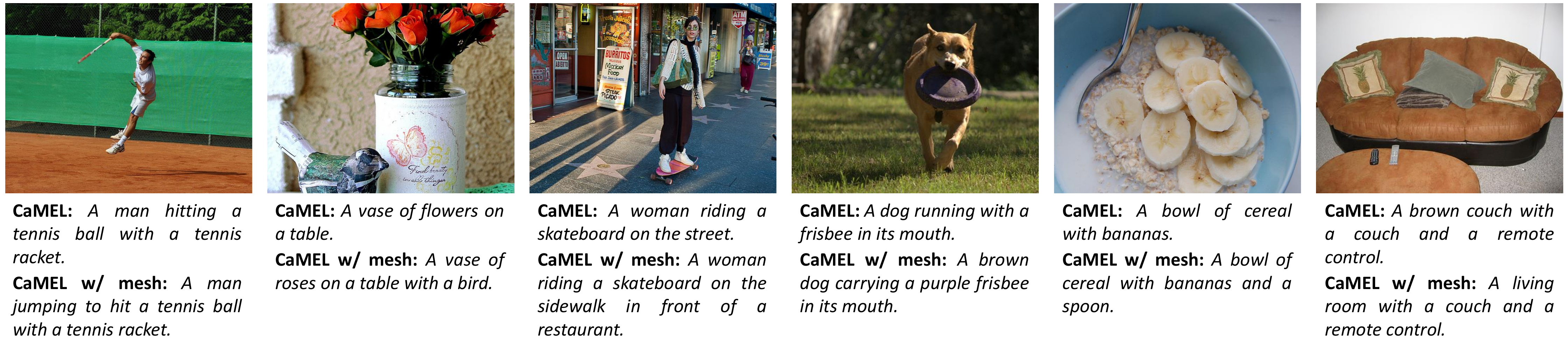}
\vspace{-0.15cm}
\caption{Qualitative results on sample images from the COCO test set.}
\label{fig:results}
\vspace{-0.2cm}
\end{figure*}

\subsection{Ablation Study}
\tinytit{Visual features assessment} We firstly discuss the role of visual features, by comparing traditional detection-based features with grid-based features extracted from modern multi-modal models. In particular, we consider object detection features extracted from a Faster R-CNN model~\cite{anderson2018bottom} pre-trained on the Visual Genome dataset~\cite{krishnavisualgenome}, and grid-based features extracted from CLIP~\cite{radford2021learning}, which has been trained with language supervision. Since CLIP visual encoders can be either based on ViT-like or CNN-like architectures, we either employ the output of the last Transformer layer, removing the CLS token, or extract the grid of features produced immediately after the last convolutional layer.

As it can be seen from Table~\ref{tab:features}, Faster R-CNN features can be surpassed by modern multi-modal architectures, which brings a significant advantage in terms of all captioning metrics. While CLIP-RN50 performs worse than Faster R-CNN features when training a single language model without \ours and mesh-like connectivity, ViT-based and larger CNN-based models achieve better performance. The best performance is reached by the CLIP-RN50$\times$16 variant, which employs an EfficientNet-style architecture scaling. Overall, this brings an improvement of 7.1 CIDEr points with respect to traditional detection-based features (from 113.9 to 121.0) when training a single language model without the \ours technique and without mesh-like connectivity.

\begin{table}[t]
\small
\centering
\caption{Results on the COCO Karpathy-test split with different knowledge distillation strategies during CIDEr optimization.}
\label{tab:reinforcement}
\setlength{\tabcolsep}{.35em}
\resizebox{\linewidth}{!}{
\begin{tabular}{lccccccc}
\toprule
 & & B-1 & B-4 & M & R & C & S \\
\midrule
\ours\textsubscript{best} -- CLIP Text Embeddings & &  82.6 &  40.7 &  \textbf{30.3} & \textbf{60.1} & 138.2 & 24.3 \\
\midrule
\ours\textsubscript{all} -- Hungarian Matching & &  82.7 &  40.8 &  30.2 &  60.0 & 138.4 & 24.0 \\
\ours\textsubscript{best} -- Hungarian Matching & &  82.4 &  40.4 &  30.1 &  59.8 & 138.6 & 23.8 \\
\midrule
\ours\textsubscript{all} & &  \textbf{82.9} &  \textbf{41.0} &  30.2 & \textbf{60.1} & 138.5 & 24.0 \\
\ours\textsubscript{best} & & 82.7 & 40.9 & \textbf{30.3} & \textbf{60.1} & \textbf{138.9} & \textbf{24.5} \\
\bottomrule
\end{tabular}
}
\vspace{-0.15cm}
\end{table}

\tit{Role of \ours}
We then assess the impact of the proposed training technique during the XE pre-training stage, by also testing with different weights for the knowledge distillation loss, which we indicate with $\lambda_\text{KD}$. Results are reported in Table~\ref{tab:features}.
As it can be seen, \ours improves the performance when used with all the previously mentioned features, by a considerable margin. This confirms the appropriateness of using a mean teacher learning paradigm in image captioning. When using the RN50$\times$16 encoder, for instance, applying \ours with a distillation weight of 0.1 brings an improvement of 4.7 CIDEr points (121.0 vs 125.7). Finally, the \ours training strategy improves the performance also when using a mesh-like connectivity, from 122.6 to 125.0 CIDEr points when using $\lambda_\text{KD}$ equal to 0.01.

Overall, during XE pre-training CLIP features bring a $6.2\%$ relative
improvement with respect to traditional detection-based features, and \ours~introduces, in the best feature configuration, a relative advancement of $3.9\%$ with respect to a typical single model training.

\begin{table}[t]
\small
\centering
\caption{Comparison with the state of the art on the Karpathy-test split.}
\label{tab:sota_results}
\setlength{\tabcolsep}{.55em}
\resizebox{0.95\linewidth}{!}{
\begin{tabular}{lccccccc}
\toprule
 & & B-1 & B-4 & M & R & C & S \\
\midrule
Up-Down~\cite{anderson2018bottom} & & 79.8 & 36.3 & 27.7 & 56.9 & 120.1 & 21.4 \\
ORT~\cite{herdade2019image} & & 80.5 & 38.6 & 28.7 & 58.4 & 128.3 & 22.6 \\
GCN-LSTM~\cite{yao2018exploring} & & 80.9 & 38.3 & 28.6 & 58.5 & 128.7 & 22.1 \\
SGAE~\cite{yang2019auto} & & 81.0 & 39.0 & 28.4 & 58.9 & 129.1 & 22.2 \\ 
CPTR~\cite{liu2021cptr} & & 81.7 & 40.0 & 29.1 & 59.4 & 129.4 & - \\
MT~\cite{shi2020improving} & &  80.8 & 38.9 & 28.8 & 58.7 & 129.6 & 22.3 \\
AoANet~\cite{huang2019attention} & & 80.2 & 38.9 & 29.2 & 58.8 & 129.8 & 22.4 \\
$\mathcal{M}^2$ Transformer~\cite{cornia2020meshed} & & 80.8 & 39.1 & 29.2 & 58.6 & 131.2 & 22.6 \\
X-LAN~\cite{pan2020x} & &  80.8 & 39.5 & 29.5 & 59.2 & 132.0 & 23.4 \\
TCTS~\cite{huang2020teacher} & & 81.2 & 40.1 & 29.5 & 59.3 & 132.3 & 23.5 \\
X-Transformer~\cite{pan2020x} & & 80.9 & 39.7 & 29.5 & 59.1 & 132.8 & 23.4 \\
DPA~\cite{liu2020prophet} & & 80.3 & 40.5 & 29.6 & 59.2 & 133.4 & 23.3 \\
DLCT~\cite{luo2021dual} & & 81.4 & 39.8 & 29.5 & 59.1 & 133.8 & 23.0 \\
RSTNet~\cite{zhang2021rstnet} & & 81.8 & 40.1 & 29.8 & 59.5 & 135.6 & 23.3 \\
\midrule
\textbf{\ours} & & 82.7 & 40.9 & \textbf{30.3} & \textbf{60.1} & 138.9 &  \textbf{24.5} \\
\textbf{\ours w/ mesh} & & \textbf{82.8} & \textbf{41.3} & 30.2 & \textbf{60.1} & \textbf{140.6} & 23.9 \\
\midrule
VinVL$_B$~\cite{zhang2021vinvl} & & \textit{82.0} & \textit{40.9} & \textit{30.9} & \textit{60.7} & \textit{140.6} & \textit{25.1} \\
VinVL$_L$~\cite{zhang2021vinvl} & & \textit{82.0} & \textit{41.0} & \textit{31.1} & \textit{60.9} & \textit{140.9} & \textit{25.2} \\ 
\bottomrule
\end{tabular}
}
\vspace{-0.15cm}
\end{table}

\begin{table*}[t]
\small
\centering
\caption{Leaderboard of various methods on the online COCO test server. The $^\dagger$ marker indicates ensemble configurations.}
\label{tab:coco_test}
\setlength{\tabcolsep}{.55em}
\resizebox{0.95\linewidth}{!}{
\begin{tabular}{lccccccccccccccccccccc}
\toprule
 & & \multicolumn{2}{c}{BLEU-1} & & \multicolumn{2}{c}{BLEU-2} & & \multicolumn{2}{c}{BLEU-3} & & \multicolumn{2}{c}{BLEU-4} & & \multicolumn{2}{c}{METEOR} & &  \multicolumn{2}{c}{ROUGE} & & \multicolumn{2}{c}{CIDEr} \\
\cmidrule{3-4} \cmidrule{6-7} \cmidrule{9-10} \cmidrule{12-13} \cmidrule{15-16} \cmidrule{18-19} \cmidrule{21-22} 
& & c5 & c40 & & c5 & c40 & & c5 & c40 & & c5 & c40 & & c5 & c40 & & c5 & c40 & & c5 & c40 \\
\midrule
Up-Down~\cite{anderson2018bottom}$^\dagger$ & & 80.2 & 95.2 & & 64.1 & 88.8 & & 49.1 & 79.4 & & 36.9 & 68.5 & & 27.6 & 36.7 & & 57.1 & 72.4 & & 117.9 & 120.5 \\ 
SGAE~\cite{yang2019auto}$^\dagger$ & & 81.0 & 95.3 & & 65.6 & 89.5 & & 50.7 & 80.4 & & 38.5 & 69.7 & & 28.2 & 37.2 & & 58.6 & 73.6 & & 123.8 & 126.5 \\
TCTS~\cite{huang2020teacher} & & 80.5 & 94.8 & & 65.3 & 89.1 & & 50.9 & 80.5 & & 39.0 & 70.3 & & 29.0 & 38.4 & & 58.9 & 74.0 & & 125.3 & 127.2 \\
CPTR~\cite{liu2021cptr} & & 81.8 & 95.0 & & 66.5 & 89.4 & & 51.8 & 80.9 & & 39.5 & 70.8 & & 29.1 & 38.3 & & 59.2 & 74.4 & & 125.4 & 127.3 \\
AoANet~\cite{huang2019attention}$^\dagger$ & & 81.0 & 95.0 & & 65.8 & 89.6 & & 51.4 & 81.3 & & 39.4 & 71.2 & & 29.1 & 38.5 & & 58.9 & 74.5 & & 126.9 & 129.6 \\
$\mathcal{M}^2$ Transformer~\cite{cornia2020meshed}$^\dagger$ & & 81.6 & 96.0 & & 66.4 & 90.8 & & 51.8 & 82.7 & & 39.7 & 72.8 & & 29.4 & 39.0 & & 59.2 & 74.8 & & 129.3 & 132.1 \\
X-Transformer~\cite{pan2020x}$^\dagger$ & & 81.9 & 95.7 & & 66.9 & 90.5 & & 52.4 & 82.5 & & 40.3 & 72.4 & & 29.6 & 39.2 & & 59.5 & 75.0 & & 131.1 & 133.5 \\
DPA~\cite{liu2020prophet}$^\dagger$ & & 81.8 & 96.3 & & 66.5 & 91.2 & & 51.9 & 83.2 & & 39.8 & 73.3 & & 29.6 & 39.3 & & 59.4 & 75.1 & & 130.4 & 133.7 \\
RSTNet~\cite{zhang2021rstnet}$^\dagger$ & & 82.1 & 96.4 & & 67.0 & 91.3 & & 52.2 & 83.0 & & 40.0 & 73.1 & & 29.6 & 39.1 & & 59.5 & 74.6 & & 131.9 & 134.0 \\
DLCT~\cite{luo2021dual}$^\dagger$ & & 82.4 & 96.6 & & 67.4 & 91.7 & & 52.8 & 83.8 & & 40.6 & 74.0 & & 29.8 & 39.6 & & 59.8 & 75.3 & & 133.3 & 135.4 \\
\midrule
\textbf{\ours} & & 82.2 & 96.6 & & 67.2 & 91.7 & & 52.5 & 83.8 & & 40.2 & 73.7 & & 30.0 & 39.6 & & 59.6 & 75.2 & & 133.7 & 136.4 \\
\textbf{\ours w/ mesh} & & 82.6 & 96.8 & & 67.5 & 91.9 & & 52.8 & 83.9 & & 40.5 & 73.8 & & 29.9 & 39.4 & & 59.8 & 74.9 & & 135.1 & 137.7 \\
\textbf{\ours w/ mesh}$^\dagger$ & & \textbf{83.2} & \textbf{97.3} & & \textbf{68.3} & \textbf{92.7} & & \textbf{53.6} & \textbf{84.8} & & \textbf{41.2} & \textbf{74.9} & & \textbf{30.2} & \textbf{39.7} & & \textbf{60.2} & \textbf{75.6} & & \textbf{137.5} & \textbf{140.0} \\
\bottomrule
\end{tabular}
}
\vspace{-0.15cm}
\end{table*}

\tit{Evaluation of different SCST strategies}
Turning to the evaluation of the SCST fine-tuning stage, in Table~\ref{tab:reinforcement} we compare the performance of different knowledge distillation strategies. 
In particular, we experiment the \ours\textsubscript{best} version, in which we pair the two logits sequences with the highest probability from both models to compute the KD loss, and the \ours\textsubscript{all} version where we use all the sequences generated by the beam search algorithm, pairing them in sorted order of log probability.
Further, we explore the use of CLIP text embeddings in the MSE loss. In \ours\textsubscript{best} with CLIP text embeddings, we select the most probable caption in each of the two beams, and then apply the MSE loss between the CLS tokens given by the CLIP text encoder.

Moreover, we investigate the usage of the Hungarian matching algorithm~\cite{kuhn1955hungarian} to couple captions in the online and target beams, using their CLIP embedding similarity as distance function. We then compute the MSE loss on the original logits between all pairs -- in \ours\textsubscript{all} version -- or only the most probable caption from the target model and its most similar one from the online model -- in \ours\textsubscript{best} version.
The best version, based on all metrics except BLEU scores, is \ours\textsubscript{best} without additional algorithms, while for the BLEU metrics the best one is \ours\textsubscript{all} always without the use of additional algorithms to pair captions.

\subsection{Comparison with the state of the art}
We compare the results of \ours with those of several recent image captioning models trained without large-scale vision-and-language pre-training. In our analysis, we include methods with LSTM-based language models and attention over image regions such as Up-Down~\cite{anderson2018bottom}, either enhanced with graph-based encoding (\ie~GCN-LSTM~\cite{yao2018exploring}, SGAE~\cite{yang2019auto}, and MT~\cite{shi2020improving}) or self-attention (\ie~AoANet~\cite{huang2019attention}, X-LAN~\cite{pan2020x}, DPA~\cite{liu2020prophet}, and TCTS~\cite{huang2020teacher}), and captioning architectures entirely based on the Transformer network such as ORT~\cite{herdade2019image}, $\mathcal{M}^2$ Transformer~\cite{cornia2020meshed}, X-Transformer~\cite{pan2020x}, CPTR~\cite{liu2021cptr}, DLCT~\cite{luo2021dual}, and
RSTNet~\cite{zhang2021rstnet}.

\tit{Performance on COCO}
As it can be observed from Table~\ref{tab:sota_results}, our proposal reaches 138.9 CIDEr points, beating all the compared approaches. Adding the mesh-like connectivity to the decoder further improves the results to 140.6 CIDEr points. This represents an increase of 5.0 CIDEr points with respect to the current state of the art when training on the COCO dataset exclusively~\cite{zhang2021rstnet}. Further, in Fig.~\ref{fig:first_page} we compare the aforementioned approaches in terms of both CIDEr and number of parameters. As it can be noticed, not only \ours reports state-of-the-art results in terms of caption quality, but it also features a significant reduction in terms of number of trainable weights. As most of previous literature has increased caption quality by increasing the model capacity, our approach represents an outlier in this trend, and demonstrates that state-of-the-art CIDEr levels can be obtained even with a very lightweight model.

Finally, in Table~\ref{tab:sota_results} and Fig.~\ref{fig:first_page} we also report the performance obtained by a recent approach which employs large-scale pre-training on external data, \ie~VinVL~\cite{zhang2021vinvl}. While this approach is not directly comparable with \ours considering that it employs more data, we notice that our proposal is on pair with the Base version of VinVL, and only 0.3 CIDEr points below its Large version. In terms of model size and number of parameters, VinVL is also extremely more demanding than our proposal (cfr. Fig.~\ref{fig:first_page}). This further confirms that the usage of proper visual features and of mean teacher learning strategy can achieve a good caption quality with a reduced model size.

\tit{Online evaluation}
Finally, we also report the performance of our method on the online COCO test server\footnote{\scriptsize\url{https://competitions.codalab.org/competitions/3221}}. In this case, we also employ an ensemble of four models trained with the mesh-like connectivity. The evaluation is done on the COCO test split, for which ground-truth annotations are not publicly available. Results are reported in Table~\ref{tab:coco_test}, in comparison with the top-performing approaches of the leaderboard. As it can be seen, our method surpasses the current state of the art on all metrics, achieving an advancement of 4.2 CIDEr points with respect to the best performer.

\section{Conclusion}
\label{sec:conclusion}
In this work, we investigated the use of the mean teacher learning paradigm applied to the image captioning task. We presented \ours, Captioner with Mean tEacher Learning, a novel Transformer-based network that is trained with the interaction of two different language models that learn from each other through knowledge distillation and model averaging. Experimentally, we validated our method with different knowledge distillation strategies and visual feature extractors, surpassing the current state of the art on the COCO dataset without using external data and pre-training strategies. Furthermore, \ours~achieves similar performance to other recent proposals that make use of large-scale pre-training, while being much smaller in terms of number of parameters.


\section*{Acknowledgment}
This work has been supported by ``Fondazione di Modena'', by the ``Artificial Intelligence for Cultural Heritage (AI4CH)'' project, co-funded by the Italian Ministry of Foreign Affairs and International Cooperation, and by the H2020 ICT-48-2020 HumanE-AI-NET and H2020 MSCA ``PERSEO - European Training Network on PErsonalized Robotics as SErvice Oriented applications'' projects.

\bibliographystyle{IEEEtran}
\bibliography{bibliography}
%
%
%

\end{document}